\documentclass{article}

\usepackage{arxiv}
\usepackage[utf8]{inputenc} 
\usepackage[T1]{fontenc}    
\usepackage{hyperref}       
\usepackage{url}            
\usepackage{booktabs}       
\usepackage{amsfonts}       
\usepackage{nicefrac}       
\usepackage{microtype}      
\usepackage{lipsum}		
\usepackage{graphicx}
\usepackage[numbers]{natbib}
\usepackage{doi}
\usepackage{amsmath}
\usepackage{makecell}
\usepackage{booktabs}
\usepackage[table]{xcolor}
\usepackage{caption}
\usepackage{listings}
\usepackage{xcolor}
\usepackage{array}

\title{A Geometrical Approach to Constructing Neural Networks for Perfect Training-set Classification}

\author{ \href{https://orcid.org/0000-0001-6410-8895}{\includegraphics[scale=0.06]{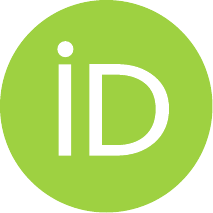}\hspace{1mm}\'Akos Hajnal}\textsuperscript{a,b}\\
	\textsuperscript{a}Laboratory of Parallel and Distributed Systems\\
        Institute for Computer Science and Control (SZTAKI) \\
        \textsuperscript{b}John von Neumann Faculty of Informatics \\
        Obuda University \\
	Budapest, Hungary \\
	\texttt{akos.hajnal@sztaki.hu} \\
}

\renewcommand{\undertitle}{} 
\renewcommand{\shorttitle}{\textit{arXiv}}

\hypersetup{
pdftitle={A template for the arxiv style},
pdfsubject={q-bio.NC, q-bio.QM},
pdfauthor={David S.~Hippocampus, Elias D.~Striatum},
pdfkeywords={First keyword, Second keyword, More},
}

\begin{document}
\maketitle

\begin{abstract}
This paper proposes a method for constructing a neural network that achieves perfect classification of all samples in a given training set, without taking regularization into account. The network is constructed on-demand, with its layer sizes and neuron parameters determined through a geometrical analysis of the input data, in which separating hyperplanes are created and mapped to neurons with step activation functions. The proposed architecture consists of two hidden layers with distinct roles: inter-class separation and transformation of the data into a linearly separable representation. Inter-class separation requires every bi-chromatic pair of samples belonging to different classes to be separated by at least one hyperplane. We show that when this separation is achieved through a hierarchical partitioning of the data, the leaf nodes of the oblique decision tree, obtained as a byproduct, directly define the neurons of the linear separability layer, whose size is only one neuron greater than that of the inter-class separation layer. Experiments on the MNIST dataset demonstrate that the proposed method achieves perfect training-set classification, and compact models. The direct correspondence between the decision tree and network neurons enhances interpretability and explainability, while the construction procedure offers opportunities for parallel implementation and improved scalability.

\end{abstract}



\section{Introduction}

This paper concerns with creating dense, feedforward, multi-layer neural networks, with the objective to achieve perfect classification of the training set. Regularization is out of the scope; no attempt is made to discover generalizable regularities or latent correlations in the data, to achieve high accuracy on the test set.

Rather than training the parameters of a predefined architecture, the proposed approach creates the layers and neurons on-demand based on a geometrical interpretation of the data (points) and of neurons (hyperplanes). The use of the step function as activation function distills the operation of neurons to linear separation, and assigns distinct roles to the individual layers; moreover, the proposed method enables highly parallelizable model construction, can produce remarkably compact networks, and enhance interpretability through its relationship with decision trees.

For classification tasks, the paper proposes a three-layer neural network architecture comprising: (1) an inter-class separation layer, (2) a class-wise linear separability layer, and (3) a decision layer. 

The objective of the first hidden layer is to separate every pair of input samples belonging to different classes. For this reason, a set of hyperplanes is constructed such that at least one hyperplane separates every \textit{bi-chromatic} pair of input points. Each hyperplane $w^Tx+b=0$ is then associated with a neuron whose weight vector and bias are given by normal vector $w$ and bias $b$, respectively, which collectively form the first layer. Using the step function as an activation function, the first layer thus maps the real-valued feature vectors of the input samples to binary vectors, while preserving their class labels.

The objective of the second hidden layer is to further transform this binary representation into a dataset in which the different classes are linearly separable. Geometrically, this transformation maps the set of labeled vertices of a unit hypercube onto another unit hypercube, potentially of a different dimension, in which the vertices corresponding to each class can be linearly separated from those belonging to other classes.

The third layer makes the final classification decision, using one neuron per class to separate that class from all others with a hyperplane representing the linear decision boundary.

To demonstrate that such three-layer architecture is capable of perfectly classifying any dataset, we present a naive construction. We assume that the input dataset contains no conflicting samples; that is, identical feature vectors are not assigned to different labels.

First, consider binary classification with positive and negative class labels, and let $P$ and $N$ denote the sets of positive and negative samples, respectively. The neurons of the first layer are obtained by creating one hyperplane for every distinct bi-chromatic pair $(p,n)$, where $p\in P$ and $n\in N$, which separates $p$ and $n$ (e.g., perpendicular bisector). The neurons of the second layer are obtained by creating one hyperplane for every positive sample of the transformed dataset, that isolates that positive hypercube vertex, being the only one on its positive side. In this way, all negative vertices are mapped to the origin by the second layer. The single output neuron then performs the final linear separation by using a hyperplane that places only the origin on the negative side, effectively implementing an OR operation. This construction achieves perfect classification with $\lvert P\rvert\lvert N\rvert+\lvert P\rvert+1$ neurons.


In multi-class classification, first-layer neurons are constructed from hyperplanes that separate every pair of samples belonging to different classes. The second-layer neurons individually isolate the labeled vertices. In the decision layer, one neuron is associated with each class and connected only to the second-layer neurons corresponding to its respective class and implementing an OR operation. This network yields perfect, one-hot-encoded predictions on the training set. The resulting model is, however, also quadratic in size with respect to the number of samples.

Such models are impractical for real-world datasets due to their prohibitive space requirements, and although determining the minimum number of required neurons, i.e. the optimal network, is an NP-hard problem, this paper presents a heuristics to achieve perfect training-set classification with a moderate number of neurons.  

Such models are impractical for real-world datasets due to their prohibitive memory requirements. Although finding the \textit{optimal} network with the minimum number of neurons is an NP-hard problem, this paper proposes heuristics for achieving perfect training-set classification with a moderate number of neurons.


The main contributions of this paper are:
\begin{itemize}
\item proposal of a two-hidden layer neural network capable of perfectly classifying arbitrary finite, non-conflicting training sets;
\item a geometrical construction method that determines the network architecture and neuron parameters directly from the training data;
\item efficiently parallelizable technique for constructing substantially more compact networks than the naive construction.
\end{itemize}

The remainder of the paper is organized as follows. Section 2 presents heuristics for constructing the proposed three-layer neural network for binary classification and evaluates them empirically on the MNIST dataset using an ensemble of binary classifiers. Section 3 extends the method to multi-class classification and evaluates it on the same dataset using a single connected model. Section 4 reviews related work and compares it with our approach. Finally, Section 5 concludes the paper.

\section{Binary classification}

In binary classification, the data are divided into two classes, referred to as \textit{positives} and \textit{negatives}. In this section, we propose a heuristic for constructing the neurons of each layer.

\subsection{Inter-class separation layer}

The objective of the first layer is to separate every bi-chromatic pair of samples using preferably the minimum number of hyperplanes. Finding such a minimum set is NP-hard in general, as the problem can be formulated as a Set Cover problem. 

For a given hyperplane, the number of bi-chromatic edges it cuts is given by $TP\cdot TN + FP\cdot FN$, where $TP$ and $TN$ denote the numbers of positive and negative samples lying on their respective sides of the hyperplane, while $FP$ and $FN$ denote the numbers of negative and positive samples lying on the opposite, incorrect sides, respectively. If the set of samples is linearly separable, this score attains its maximum possible value when the hyperplane correctly classifies the entire set. Since even \textit{testing} whether a dataset is linearly separable can be computationally expensive \cite{hajnal2026linear}, instead of finding the hyperplane that yields the maximum bi-chromatic edge cut, we only aim to maximize the number of correct classifications ($TP+TN$). 

For this purpose, we use established classification methods, such as linear discriminant analysis, logistic regression, linear-kernel support vector machines, or the Perceptron algorithm, and follow a divide-and-conquer strategy. The proposed heuristic recursively partitions the dataset by splitting each subset along a hyperplane determined by the selected classification method, until all resulting subsets are homogeneous. This procedure ensures that every bi-chromatic pair of samples is eventually separated by at least one hyperplane. 

Given the separating hyperplane obtained at each split, we perform an additional optimization step: we fine-tune its bias to maximize $TP\cdot TN + FP\cdot FN$, while keeping the hyperplane's orientation unchanged.

Despite its simplicity, the proposed technique can significantly reduce the number of neurons required, as demonstrated in Section \ref{bin-res}: for the MNIST dataset, it yields networks with approximately 11--91 first-layer neurons, compared with more than 300,000 neurons that would be required by the naive approach.

\subsection{Class-wise linear separability layer}

The objective of the second layer is to further transform the binary representation produced by the first layer into a dataset in which, for each class, its samples are linearly separable from all samples belonging to the other classes. As noted earlier, even finding the minimum number of neurons to achieve bi-chromatic separation is NP-hard in general (bi-chromatic separation in the second layer is also necessary), requiring the further transformed dataset to be linearly separable makes the problem even more difficult \cite{baum1988capabilities}.

However, rather than treating the binary dataset as an independent input to the second layer, we can exploit the hierarchical structure of the data induced by the divide-and-conquer procedure used to construct the first-layer neurons. 

We simultaneously build a decision tree, while constructing the first-layer neurons, in which each decision node corresponds to a selected separating hyperplane. Accordingly, the False and True branches correspond directly to first-layer neuron outputs of $0$ and $1$, respectively, for a given sample, due to the step activation function. Each leaf node corresponds to a homogeneous subset of samples belonging to a single class, all of which satisfy the same sequence of decision outcomes along the path to that leaf. These path conditions can be encoded as patterns of first-layer neuron outputs.

The second-layer neurons are thus constructed by iterating over the positive leaf nodes and creating one neuron for each leaf. For each neuron, we determine the path conditions leading to the related leaf and assign connection weights to the first-layer neurons as follows: a weight of $1$ if the corresponding decision along the path is True, $-1$ if it is False, and $0$ if the corresponding decision does not occur along the path.

The bias is set to $0.5$ minus the number of weights equal to $1$. Consequently, for any positive sample belonging to a positive leaf, all path conditions are satisfied, yielding a value of $0.5$ and an output of $1$ after applying the step activation function. Samples belonging to other leaves violate at least one of these conditions, resulting in a negative value and an output of $0$. Thus, each positive sample activates exactly one neuron in the second layer, whereas negative samples produce an all-zero activation pattern, thereby, mapping all negative samples, and only negative samples, to the origin, resulting in a linearly separable representation.


\subsection{Decision layer}

The final decision is made by a single neuron in the decision layer using step activation function, which implements an OR operation over the second-layer neurons, with all connection weights set to $1$ and bias set to $-0.5$.

We note that, when constructing the second layer, if the number of negative leaves turns out to be smaller than the number of positive leaves, the same procedure can instead be applied to the negative leaves, with the output of the decision-layer neuron inverted. Since a binary decision tree has one more leaf node than decision nodes, and the latter correspond to the neurons in the first layer, the second layer can constructed using at most $\lceil l_1/2\rceil$ neurons, where $l_1$ is the size of the first layer.

\subsection{Empirical results}\label{bin-res}

The proposed method was evaluated on the MNIST handwritten digits dataset \cite{lecun1998a}, which contains 60,000 training samples of $28\times28$-pixel images (784 features) representing handwritten digits from 0 to 9. We addressed the task of recognizing each digit in the training set using an ensemble of binary classifiers, each associated with a single digit and trained in a one-vs-rest manner. Each classifier has a single output indicating whether the input image belongs to the corresponding class. The binary classifiers collectively determine the prediction in a one-hot-encoded form.

We applied three different hyperplane separation techniques in our Python implementation\footnote{Source code: https://github.com/ahajnal/three-layers-are-all-you-need/blob/main/Binary.ipynb}: the Perceptron algorithm, a linear support vector classifier (SVC), and logistic regression (LR). We used the \texttt{LogisticRegression} and \texttt{SVC} classes provided by the scikit-learn machine-learning framework; the Perceptron algorithm was implemented directly based on Rosenblatt's method \cite{rosenblatt1957perceptron}. The SVC was limited to 1,000 iterations (\texttt{max\_iter=1000}), while the Perceptron algorithm was limited to 100 epochs. The reported results are from a single execution\footnote{Host machine: Intel Xeon Processor (Cascade Lake), 2.1 GHz, 32 GB RAM}; other executions yielded similar results, with little variation.

Table \ref{table1} summarizes the characteristics of the constructed networks, including the separation technique, the individual network sizes for each digit, the total model size in terms of the number of neurons, the time required to construct each binary classifier, the overall model construction time (applying sequential construction), the total number of parameters, and the overall training-set accuracy (to verify the correctness of the implementation).

The results show that all of the evaluated methods substantially reduce the number of required neurons compared to the naive approach. With these settings, logistic regression produced the smallest networks (with a total of 381 neurons), and achieved the shortest construction time (about 15 minutes).

We note that the implementation did not exploit parallelization. However, individual binary classifiers could be constructed concurrently using separate processes, while the inter-class separation procedure could benefit from parallel construction of decision-tree branches across data partitions, for example, using multiple threads. These strategies could substantially reduce the overall construction time.


\begin{table*}
\caption{Binary classifier ensembles recognizing individual digits, constructed using different hyperplane selection methods on the MNIST training dataset.}
\footnotesize
\setlength{\tabcolsep}{6pt}
\begin{tabular}{l c c c c c c}
\hline

\makecell{Method} &
\makecell{Layer\ sizes} &
\makecell{Total number\\ of neurons} &
\makecell{Construction\\ times} &
\makecell{Total construction\\ time (sequential)} &
\makecell{Total number\\ of parameters} &
\makecell{Training\\ accuracy} \\

\hline


\makecell{Perceptron Algorithm} &\makecell{0: 27-13-1\\1: 21-11-1\\2: 37-18-1\\3: 55-27-1\\4: 37-19-1\\5: 47-23-1\\6: 27-14-1\\7: 30-15-1\\8: 91-45-1\\9: 77-39-1} &\makecell{683} &\makecell{0: 134 s\\1: 107 s\\2: 148 s\\3: 163 s\\4: 124 s\\5: 151 s\\6: 120 s\\7: 118 s\\8: 171 s\\9: 155 s} &\makecell{1 389 s} &\makecell{365 366} &\makecell{100\%}  \\

\hline

\makecell{Support Vector Machine\\ (SVC)} &\makecell{0: 16-8-1\\1: 17-9-1\\2: 39-20-1\\3: 46-23-1\\4: 23-11-1\\5: 46-23-1\\6: 24-13-1\\7: 28-14-1\\8: 58-29-1\\9: 51-26-1} &\makecell{534} &\makecell{0: 154 s\\1: 143 s\\2: 234 s\\3: 245 s\\4: 182 s\\5: 242 s\\6: 167 s\\7: 183 s\\8: 261 s\\9: 243 s}  &\makecell{2 056 s} &\makecell{280 684} &\makecell{100\%}  \\

\hline

\makecell{Logistic Regression\\(LR)} &\makecell{0: 11-6-1\\1: 11-6-1\\2: 22-12-1\\3: 30-16-1\\4: 17-9-1\\5: 27-14-1\\6: 15-8-1\\7: 17-9-1\\8: 51-26-1\\9: 42-22-1} &\makecell{381} &\makecell{0: 73 s\\1: 75 s\\2: 95 s\\3: 101 s\\4: 77 s\\5: 93 s\\6: 76 s\\7: 79 s\\8: 121 s\\9: 109 s} &\makecell{899 s} &\makecell{194 951} &\makecell{100\%}  \\

\hline

\end{tabular}
\label{table1}
\end{table*}

\section{Multi-class classification}

In multi-class classification, each sample belongs to exactly one of a finite set of more than two classes. Our objective is to construct a single connected network with multiple output neurons, each corresponding to a class and collectively producing a one-hot-encoded prediction. The approach presented in this section largely follows the binary case but extends it to accommodate multiple classes.

\subsection{Inter-class separation layer}

As in the binary case, the neurons in the first layer represent a set of hyperplanes such that every pair of samples belonging to different classes is separated by at least one hyperplane in the set. We use the same divide-and-conquer approach to construct these neurons while simultaneously building a decision tree. However, since the data contain more than two classes, the selection of separating hyperplanes differs from that in the binary case.

Our greedy strategy aims to detach the largest possible homogeneous subsets as early as possible in the partitioning process. Therefore, we consider each class in turn and seek a hyperplane that separates samples of the selected class from those belonging to all other classes (one-vs-rest). For each candidate hyperplane, we then optimize its bias to maximize the score $TP\cdot TN + FP\cdot FN$, treating samples of the selected class as positive and all other samples as negative. We select the hyperplane that achieves the highest score to partition the current subset, and recursively apply the procedure to any resulting non-homogeneous subsets.


\subsection{Class-wise separability Layer}

The neurons in the second layer, which further transform the dataset into a representation in which each class is linearly separable from all others, can be constructed by mapping each leaf node of the decision tree formed as a byproduct of the first-layer construction to a corresponding neuron in the second layer.

\begin{figure*}[t]
\centering
\includegraphics[width=0.8\linewidth]{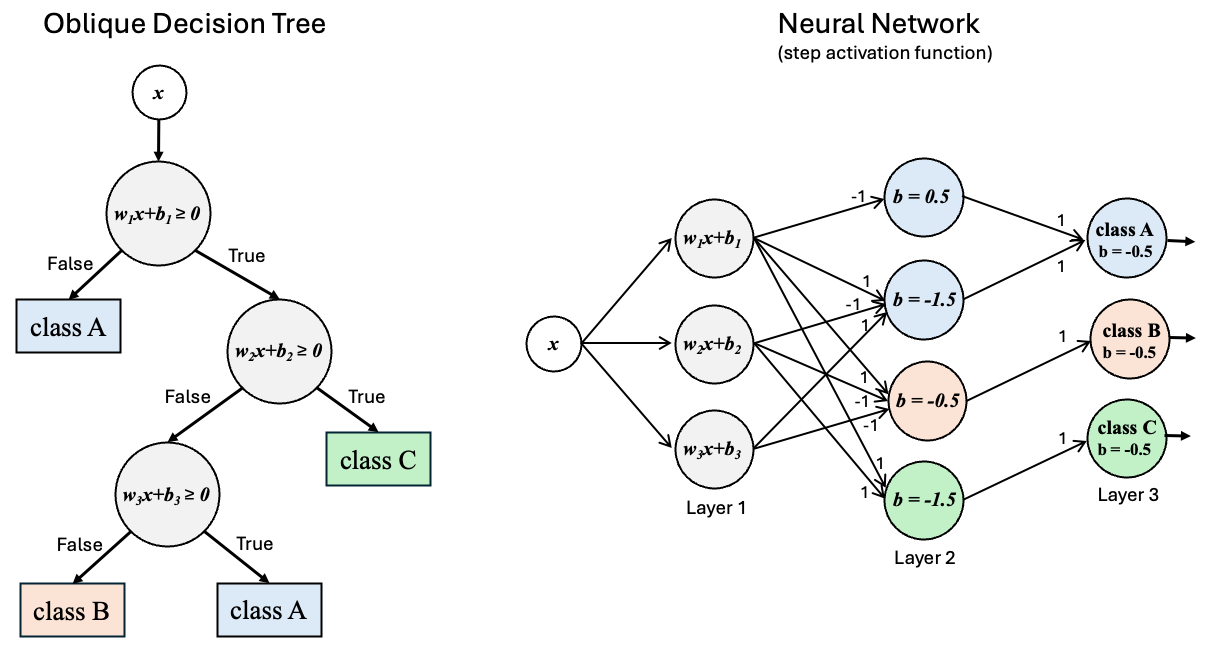}
\caption{An example of an oblique decision tree and its corresponding neural network.}\label{fig1}
\end{figure*}

Neuron weights and biases can be determined analogously to the binary case by encoding the path conditions in the weights of the connections between the first- and second-layer neurons and in the corresponding bias values. Fig. \ref{fig1} illustrates an example of an oblique decision tree (left) and the corresponding neural network (right). The weights of the connections to the second and third layers are indicated next to the edges, while the biases are shown inside the neurons. Connections with zero weights are omitted for clarity.

As in the binary case, each sample activates exactly one neuron in the second layer, and the size of the first layer corresponds to the number of decision nodes. In contrast, since multi-class labels are not complementary, the size of the second layer corresponds directly to the number of leaf nodes.

\subsection{Decision layer}

In the decision layer, one neuron is created for each class. Each output neuron is connected to the second-layer neurons corresponding to leaf nodes of that class, with connection weights set to $1$ and the bias set to $-0.5$ (all other connection weights are set to $0$). This implements an OR operation over the second-layer neurons of the respective class and ensures activation if and only if any of them is activated.

\subsection{Empirical results}

Table \ref{table2} summarizes the characteristics of the models constructed for the MNIST training dataset. This implementation\footnote{Source code: https://github.com/ahajnal/three-layers-are-all-you-need/blob/main/Multiclass.ipynb} used linear discriminant analysis (LDA), linear support vector classifier (SVC), and logistic regression (LR) to identify separating hyperplanes, and did not exploit parallelization either.

Since the ensemble of binary classifiers presented in the previous section introduces redundancy -- each bi-chromatic sample pair is separated independently by at least two of the corresponding binary classifiers -- we expected the resulting networks to be smaller. However, the results show the opposite: the constructed networks are larger; although, construction time had become shorter when using logistic regression.

\begin{table*}
\caption{Multi-class classifiers recognizing individual digits, constructed using different hyperplane selection methods on the MNIST training dataset.}
\footnotesize
\setlength{\tabcolsep}{10.5pt}
\begin{tabular}{l c c c c c c}
\hline

\makecell{Method} &
\makecell{Layer\ sizes} &
\makecell{Total number\\ of neurons} &
\makecell{Total construction\\ time (sequential)} &
\makecell{Total number\\ of parameters} &
\makecell{Training\\ accuracy} \\

\hline

\makecell{Linear Discriminant Analysis\\ (LDA)} &\makecell{625-626-10} &\makecell{1 261} &\makecell{384 s}  &\makecell{888 771} &\makecell{100\%}  \\

\makecell{Support Vector Machine\\ (SVC)} &\makecell{356-357-10} &\makecell{723} &\makecell{2 070 s}   &\makecell{410 489} &\makecell{100\%}  \\

\makecell{Logistic Regression\\(LR)} &\makecell{278-279-10} &\makecell{567} &\makecell{582 s} &\makecell{298 871} &\makecell{100\%}  \\

\hline

\end{tabular}
\label{table2}
\end{table*}

\section{Related Work}

The role of hidden layers in enhancing the linear separability of successive data representations in deep neural networks has also been recognized by other authors (e.g., \cite{alain2016understanding, apicella2024hidden}). This insight motivated the design and the distinct roles assigned to the two hidden layers in our proposed architecture.

Our approach to achieving linear separability proceeds through an intermediate structure: a decision tree obtained by geometrically partitioning the input data, which is then mapped to a neural network. After developing our method, we discovered that Sethi \cite{sethi1990entropy} had already described essentially the same mapping from decision trees to neural networks, using layers with similar functions but different names: \textit{partitioning}, \textit{ANDing}, and \textit{ORing}. Although achieving linear separability was not explicitly identified as an objective in that paper, it is an implicit consequence of the method -- when the decision tree perfectly classifies all input samples, which, in contrast, was the primary objective of our approach.

The hierarchical partitioning of samples performed by our first layer is closely related to decision tree construction: the fewer decision nodes required, the fewer neurons are needed in the resulting network. Conventional decision tree algorithms typically select splits based on entropy reduction or Gini impurity. In contrast, our heuristic employs classification methods to select hyperplanes that maximize the number of bi-chromatic edges cut. Although this criterion is related to impurity-based splitting, it is not equivalent to entropy reduction or Gini impurity minimization.

The multivariate partitioning of samples using separating hyperplanes, as employed in our approach, leads to the construction of oblique decision trees. Finding an optimal oblique decision tree can be formulated as a mixed-integer linear programming problem \cite{bertsimas2017optimal}, and several other algorithms and heuristics exist for direct construction (e.g., \cite{murthy1993oc1, loh2011classification, li2025lht}). These methods could serve as alternatives to our solution.

The paper presents an empirical evaluation on the MNIST dataset, demonstrating the practical applicability of the proposed method. A related geometrical approach based on the Perceptron algorithm was presented in \cite{hajnal2022multi}, achieving perfect training-set classification using an ensemble  of binary classifiers with a single hidden layer (30 hidden neurons and one output neuron per digit, a total of $310$ neurons). That method, however, requires the network size to be specified in advance and trained the neuron weights iteratively. In contrast, the current method constructs the network on-demand as the data are analyzed, eliminating the need to specify the network size beforehand.


 
\section{Conclusions}

This paper presents a method for constructing neural networks designed to achieve perfect classification of the training data. It adopts a geometrical approach to determine neuron weights and biases directly from hyperplanes that separate data samples. In contrast to conventional approaches, the network architecture is constructed on-demand, eliminating the need to specify the architecture in advance. The correspondence between neurons and the intermediate decision tree enhances the model's interpretability and explainability.

Empirical results on the MNIST dataset demonstrate that the proposed method achieves perfect training-set classification while producing compact networks. Although the prototype implementation did not exploit parallelization, the construction offers opportunities for both data and model parallelism, potentially reducing construction time and improving scalability.

We used classical classification methods to determine the separating hyperplanes at the decision nodes, which, however, do not directly maximize the number of bi-chromatic edges cut; furthermore, our technique optimizes one decision node at a time, whereas recent studies suggest that jointly optimizing multiple decisions can lead to smaller decision trees \cite{bertsimas2017optimal}. Exploring these options could further reduce the size of the resulting models and is left for future work.

\section*{Acknowledgement}

We thank the HUN-REN Cloud \cite{heder2022past} for providing the computational resources that enabled us to carry out experiments. This research was not funded by any grant.

\section*{Dataset} The MNIST handwritten digits dataset, among other public sources,  is accessible from within the TensorFlow-Keras framework (\texttt{tensorflow.keras.datasets.mnist}).

For reproducibility of the results the source codes of the experiments are made publicly accessible in a GitHub repository: \href{https://github.com/ahajnal/three-layers-are-all-you-need}{https://github.com/ahajnal/three-layers-are-all-you-need}. 

\section*{Declaration interests}
The author declares that he has no known competing financial interests or personal relationships that could have appeared to influence the work reported in this paper.

\section*{Declaration of generative AI and AI-assisted technologies in the manuscript preparation process}
During the preparation of this work, the authors used OpenAI's GPT-5.5 large language model to improve the wording and clarity of the manuscript. After using this service, the authors carefully reviewed and edited the content as needed and take full responsibility for all aspects of the published article.

\vfill 

\bibliographystyle{IEEEtran}

\bibliography{references}  
\vfill 

\end{document}